\documentclass[10pt,twocolumn,letterpaper]{article}

\usepackage{cvpr}
\usepackage{times}
\usepackage{epsfig}
\usepackage{graphicx}
\usepackage{amsmath}
\usepackage{amssymb}
\usepackage{multirow}
\usepackage[misc]{ifsym}

\usepackage[breaklinks=true,bookmarks=false]{hyperref}

\cvprfinalcopy 

\def\cvprPaperID{8} 

\ifcvprfinal\fi
\begin{document}

\title{Multi-Domain Learning and Identity Mining for Vehicle Re-Identification}
\newcommand*{\affaddr}[1]{#1}
\newcommand*{\affmark}[1][*]{\textsuperscript{#1}}
\author{Shuting He\affmark[1,2]\thanks{The work was done when Shuting He was intern at Alibaba Group.},  Hao Luo\affmark[1,2],  Weihua Chen\affmark[2], Miao Zhang\affmark[1], Yuqi Zhang\affmark[2] \\
\vspace{0.5em}
Fan Wang\affmark[2],  Hao Li\affmark[2], Wei Jiang\affmark[1]\\
\affaddr{\affmark[1]Zhejiang University}\\
\affaddr{\affmark[2]Alibaba Group} \\
{  \tt\small shuting\_he@zju.edu.cn }
{  \tt\small haoluocsc@zju.edu.cn }
{  \tt\small kugang.cwh@alibaba-inc.com }
}

\maketitle
\thispagestyle{empty} 

\begin{abstract}
This paper introduces our solution for the Track2 in AI City Challenge 2020 (AICITY20). The Track2 is a vehicle re-identification (ReID) task with both the real-world data and synthetic data.

Our solution is based on a strong baseline with bag of tricks (BoT-BS) proposed in person ReID. At first, we propose a multi-domain learning method to joint the real-world and synthetic data to train the model. Then, we propose the Identity Mining method to automatically generate pseudo labels for a part of the testing data, which is better than the k-means clustering. The tracklet-level re-ranking strategy with weighted features is also used to post-process the results. Finally, with multiple-model ensemble, our method achieves 0.7322 in the mAP score which yields third place in the competition. The codes are available at \url{https://github.com/heshuting555/AICITY2020_DMT_VehicleReID}.

\end{abstract}

\section{Introduction}

AI City Challenge is a workshop in CVPR2020 Conference. It focuses on different computer vision tasks to make cities' transportation systems smarter. This paper introduces our solutions for the Track2, namely city-scale multi-camera vehicle re-Identification (ReID).

Vehicle ReID is an important task in computer vision. It aims to identify the target vehicle in images or videos across different cameras, especially without knowing the license plate information. Vehicle ReID is important for intelligent transportation systems (ITS) of the smart city. For instance, the technology can track the trajectory of the target vehicle and detect traffic anomalies. Recently, most of works have been based on deep learning methods in vehicle ReID, and these methods have achieved great performance on some benchmarks such as Veri-776\cite{liu2016large} and VehicleID\cite{liu2016deep}.

\begin{figure}[htb]
\centering
\includegraphics[width=0.85\linewidth]{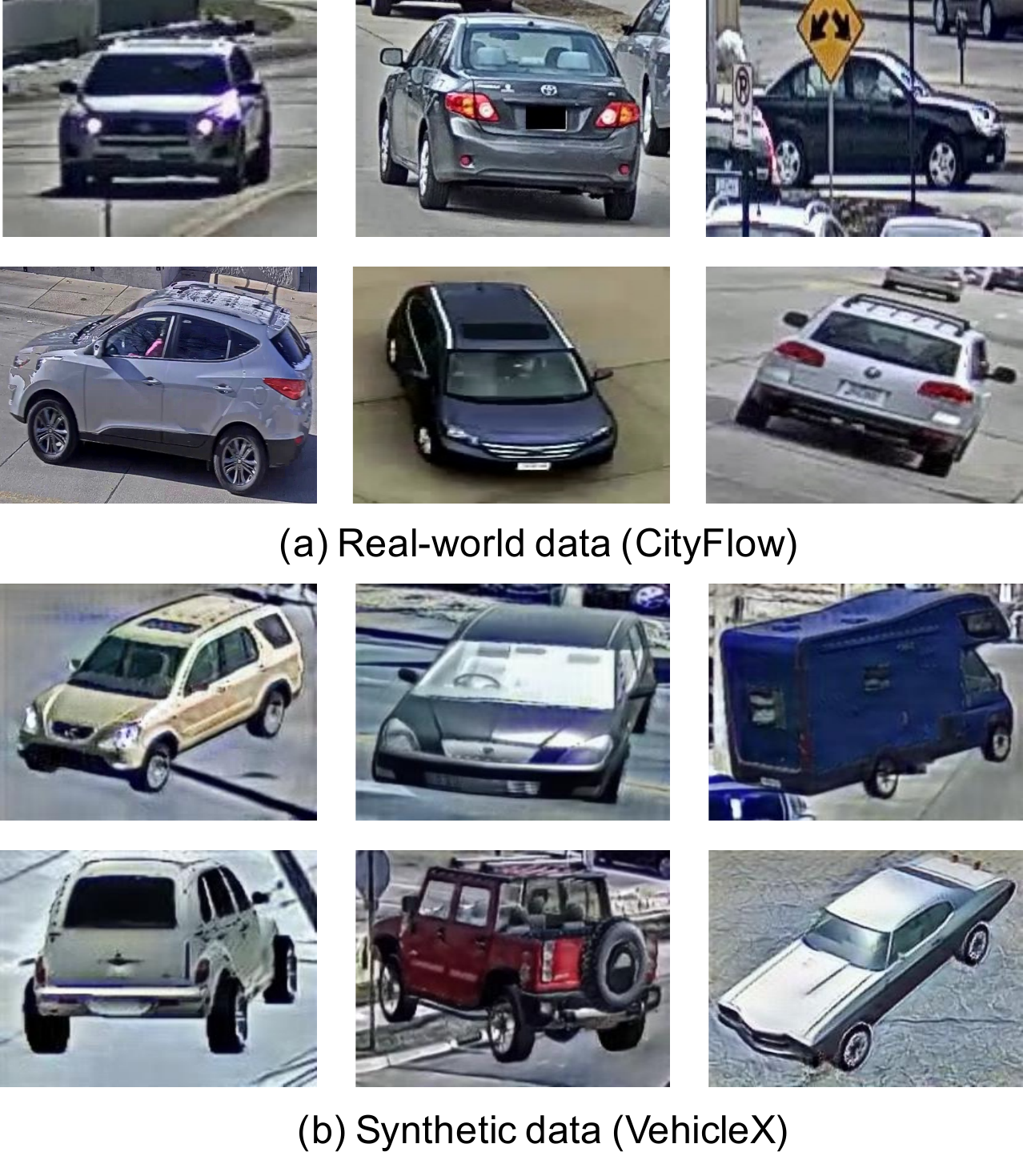}
\caption{Some examples of the real-world and synthetic data.}
\label{fig:demo}
\end{figure}

Track2 cannot be fully considered as a standard vehicle ReID task where the model is trained and evaluated on same-domain data. As shown in Figure \ref{fig:demo}, both the real-world and synthetic datasets are provided to train the model in Track2. There exists great bias between these two different datasets, so it remains some challenge that how to reasonably use the synthetic dataset. In addition, some special rules in Track2 are introduced as follow:
\begin{itemize}
\item External data cannot be used. Public person- and vehicle-based datasets, which includes Market1501 \cite{zheng2015scalable}, DukeMTMC-reID \cite{ristani2016MTMC}, VeRi , VehicleID, etc, are forbidden. Private datasets collected by other institutions are also not allowed.
\item Additional manual annotations can only be made on the training sets. Automatic annotations can be applied to the test sets also.
\item Teams can use open-source pre-trained models trained on external datasets (such as ImageNet \cite{deng2009imagenet} and MS COCO \cite{lin2014microsoft}) that have been published by other people.
\item There is no restriction on the use of synthetic data.
\item Top-100 mean Average Precision (mAP) determines the leaderboard.
\end{itemize}

Since the task is similar to person ReID, we used a strong baseline with bag of tricks (BoT-BS) in person ReID \cite{luo2019bag,luo2019strong} as the baseline in this paper. BoT-BS introduces the BNNeck to reduce the inconsistency between ID loss (cross-entropy loss) and triplet loss in the training stage. BoT-BS is not only  a strong baseline for person ReID, but also suitable for vehicle ReID while it can achieve 95.8\% rank-1 and 79.9\% mAP (ResNet50 backbone) accuracy on Veri-776 benchmark. In the paper, we modify some training settings, such as learning rates, optimizer and loss functions, etc, to improve the ReID performance on the new dataset.

Except modifying the BoT-BS, we also focus on how to use the synthetic data to improve the ReID performance on the real-world data. Because there exists great bias between the real-world and synthetic data, it is a more challenging task than cross-domain or domain adaption tasks in vehicle ReID. We observe that both directly merging the real-world and synthetic data to train models and pre-training models on the synthetic data cannot improve the ReID performance. Based on the motivation that low-level features such as color and texture are shared in the real-world and synthetic domain, we propose a multi-domain learning (MDL) method that the model is pre-trained on the real-world and a part of synthetic data and then is fine-tuned on the real-world data with the first few layers being frozen. 

In addition, the testing set is allowed to used without manual annotations. Some unsupervised methods, such as the k-means clustering, can be adopted to generate pseudo labels for the testing data. However, the pseudo labels are not accurate enough because of the poor performance of ReID models.Therefore, We propose the Identity Mining (IM) method to generate more accurate pseudo labels. IM selects some samples of different IDs with high confidence as clustering centers, where only one sample of each identity can be selected. Then, for each clustering center, some similar samples are labeled with the same ID. Different from the k-means clustering dividing all data into several clusters, our IM method just automatically labeled a part of data with high confidence.

To further improve the ReID performance, some effective methods are introduced. For instance, re-ranking (RR) strategy \cite{zhong2017re} is a widely used method to post-process the results. The original RR is a image-to-image ReID method, but the tracklet information is provided in Track2. Therefore, we introduce a tracklet-level re-ranking strategy with weighted features (WF-TRR) \cite{he2019multi}. Although our single model can reach 68.5\% mAP accuracy on CityFlow\cite{Naphade19AIC19}, we further improve the mAP accuracy to 73.2\% with the multiple-model ensemble.

Our contributions can be summarized as follow:
\begin{itemize}
\item A multi-domain learning strategy is proposed to jointly exploit the real-world and synthetic data.
\item The Identity Mining method is proposed to automatically generate pseudo labels for a part of the testing data.
\item We achieves 0.7322 in the mAP score which yields third place in the competition.
\end{itemize}

\section{Related Works}

We introduce deep ReID and some works of AICITY2019 in this section.

\subsection{Deep ReID}
Re-identification (ReID) is widely studied in the field of computer vision. This task possesses various important applications. Most existing ReID methods based on deep learning. Recently, CNN-based features have achieved great progress on both person ReID and vehicle ReID. Person ReID provides a lot of insights for vehicle ReID. Our method is based on a strong baseline \cite{luo2019bag,luo2019strong} in person ReID. For vehicle ReID, Liu \emph{et al.} \cite{liu2016deep} introduced a pipeline that uses deep relative distance learning (DRDL) to project vehicle images into an Euclidean space, where the distance can directly measure the similarity of two vehicle images. Shen \emph{et al.} \cite{shen2017learning} proposed a two-stage framework that incorporates complex spatial-temporal information of vehicles to effectively regularize ReID results. Zhou \emph{et al.} \cite{zhou2018aware} designed a viewpoint-aware attentive multi-view inference (VAMI) model that only requires visual information to solve multi-view vehicle ReID problems. While He \emph{et al.} \cite{he2019part} proposed a simple yet efficient part-regularized discriminative feature-preserving method, which enhances the perceptive capability of subtle discrepancies, and reported promising improvement. Some works also studied discriminative part-level features for better performance. Some works \cite{wang2017orientation,khorramshahi2019attention,khorramshahi2019dual} in vehicle ReID had utilized vehicle key points to learn local region features. Several recent works \cite{he2019part, wang2019vehicle,guo2019two,teng2018scan} in vehicle ReID had stated that specific parts such as windscreen, lights and vehicle brand tend to have much discriminative information. In \cite{zhu2019vehicle}, different directional part features were
utilized for spatial normalization and concatenation to serve as a directional deep learning feature for vehicle Re-ID. Qian \emph{et al.} \cite{qian2019stripe} proposed Stripe-based and Attribute-aware Network (SAN) to fuse the information of global features, part features and attributed features.

\begin{figure*}[htb]
\centering
\includegraphics[width=1.\linewidth]{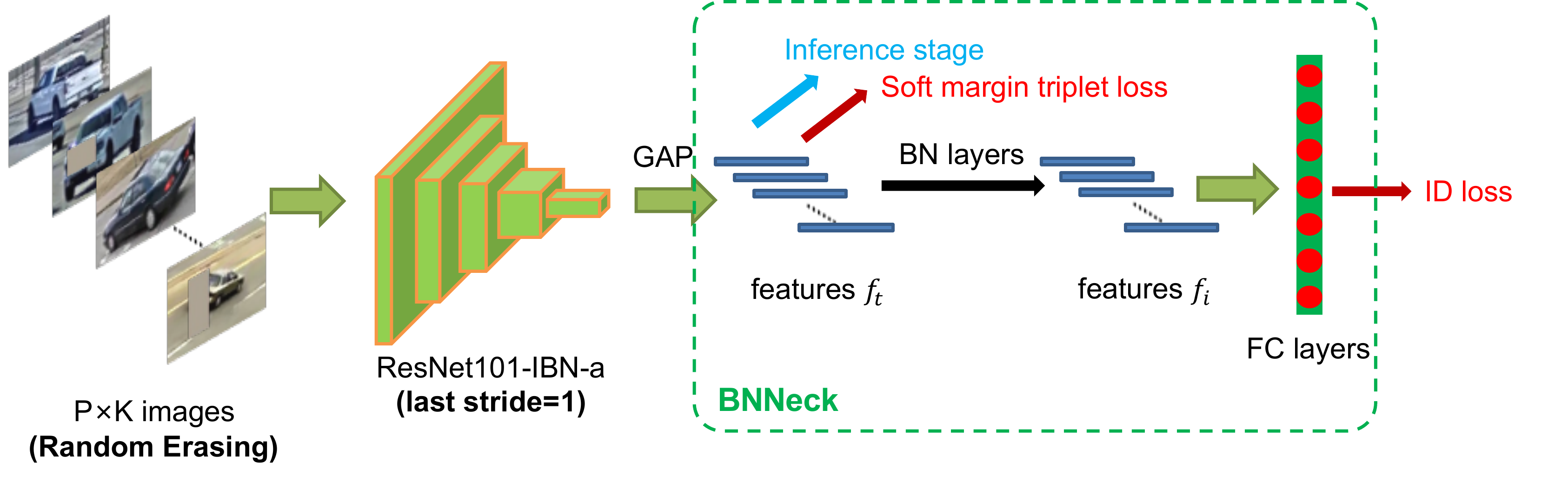}
\caption{The framework of our baseline model BoT-BS.}
\label{fig:baseline}
\end{figure*}

\subsection{AICITY19}

Since AICITY20 is updated from AI CITY Challenge 2019 (AICITY19), some methods of AICITY19 are helpful for our solution. The organizers outlined the methods of leading teams in \cite{naphade20192019}. Since some external data could be used in AICITY19, there were many different ideas last year. Tan \emph{et al.} \cite{tan2019multi} used a method based on extracting visual features from convolutional neural networks (CNNs), and leveraged semantic features from traveling direction and vehicle type classification. Huang \emph{et al.} \cite{huang2019multi} exploited vehicles semantic attributes to jointly train the model with ID labels. In addition, some leading teams were used pre-trained models to extract vehicle pose, which could infer the orientation information \cite{huang2019multi,khorramshahi2019attention}. Re-ranking methods were used as a post-processing method to improve the ReID performance by many teams\cite{huang2019multi,khorramshahi2019attention,huang2019deep,shankar2019comparative}. External data and additional annotations were add into training model by some leading teams, but it is not allowed in AICITY20.

\section{Methods}

\subsection{Baseline Model}

Baseline model is important for the final ranking. In track2, we use a strong baseline (BoT-BS) \cite{luo2019bag,luo2019strong} proposed for person ReID as the baseline model. To improve the performance on Track2' datasets, we modify some settings of BOT-BS. The output feature is followed by the BNNeck \cite{luo2019bag,luo2019strong} structure, which separates ID loss (cross-entropy loss) and triplet loss \cite{hermans2017defense} into two different embedding spaces. The triplet loss is the soft-margin version as follow:
\begin{equation}
\mathcal{L}_{Tri}=\log \left[1+ \exp(||f_a -f_p||_2^2 - ||f_a -f_n||_2^2 +m)\right]
\end{equation}
Hard example mining is used for soft-margin triplet loss. We delete center loss because it does not greatly improve the retrieval performance while increasing computing resources. We tried to modify cross-entropy loss into arcface loss, but arcface loss achieved worse performance on CityFlow. In the inference stage, we observe that the features $f_t$ obtains better performance than $f_i$ by a little margin. Due to better performance, we use the SGD optimizer instead of the Adam optimizer. To improve the performance, we train the BoS-BS with deeper backbones and larger-size images. The framework of the baseline model is shown in \ref{fig:baseline}, and more details can be referred to \cite{luo2019bag,luo2019strong}. For reference, our modified baseline achieves 96.9\% rank-1 and 82.0\% mAP accuracy on Veri-776 benchmark.



\begin{figure*}[!htb]
\centering
\includegraphics[width=1.\linewidth]{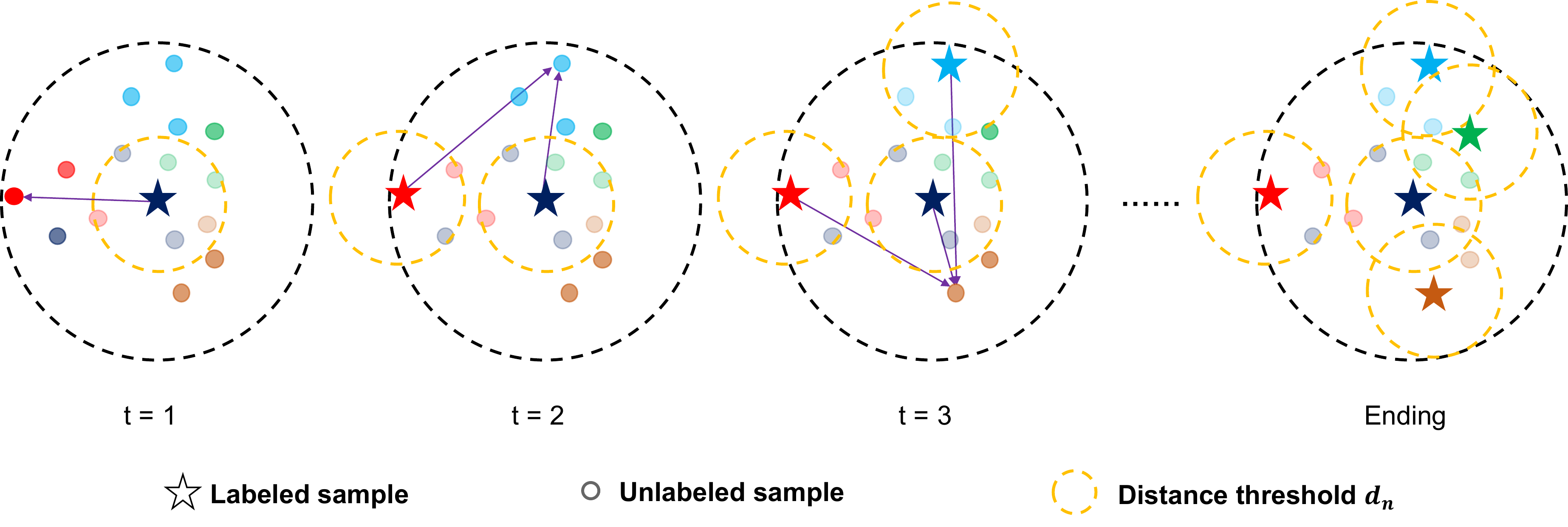}
\caption{The first stage of our Identity Mining (IM) method. Each step can label a sample with the new ID. The radius of the yellow circle is the distance threshold of the negative pair as $d_n$. The samples outside yellow circles are candidates satisfying the distance constraint. Finally,  all samples are covered by at least one yellow circle.}
\label{fig:im}
\end{figure*}

\subsection{Multi-Domain learning}
In this section, we will introduce a novel multi-domain learning (MDL) method to exploit the synthetic data. 

Both real-world and synthetic data are provided in this challenge, so how to learn discriminative features from two different domains is an important problem. For convenience, the real-world and synthetic data/domains are denoted as $\mathcal{D}_{R}$ and $\mathcal{D}_{S}$, respectively. The goal is to train a model on $\mathcal{D}_{R} \cup \mathcal{D}_{S}$ and make it achieve better performance on $\mathcal{D}_{R}$. There are two simple solutions as follow: 
\begin{itemize}
\item Solution-1: Directly merging the real-world and synthetic data to train ReID models;
\item Solution-2: Train a pre-trained model on the synthetic data $\mathcal{D}_{S}$ first and then fine-tune the pre-trained model on the real-world data $\mathcal{D}_{R}$.
\end{itemize}
However, these two solutions do not work in the challenge. Because the number of data in $\mathcal{D}_{S}$ is much larger than the one in $\mathcal{D}_{R}$, Solution-1 will cause the model to be more biased towards $\mathcal{D}_{S}$. Because there exists great bias between $\mathcal{D}_{R}$ and $\mathcal{D}_{S}$, the pre-trained model on $\mathcal{D}_{S}$ may not be better than the pre-trained model on ImageNet for CityFlow dataset. Therefore, Solution-2 is either not a good manner to solve this problem. However, some works \cite{tan2019multi, huang2019multi} used pre-trained models on Veri-776, VehicleID or CompCar\cite{yang2015large} datasets to obtain better performance in AI CITY Challenge 2019, which shows that the pre-trained model trained on reasonable data is effective. Based on aforementioned discussion, we proposed a novel MDL method to exploit the synthetic data VehicleX. The proposed method includes two stages, \emph{i.e.}, pre-trained stage and fine-tuning stage.

\textbf{Pre-trained Stage.} All training data of the real-world data $\mathcal{D}_{R}$ are denoted as the image set $\mathcal{R}$. Then, we randomly sample a part of identities from the synthetic data $\mathcal{D}_{S}$ to construct a new image set $\mathcal{S}$. The model is pre-trained on the new training set $\mathcal {R} \cup  \mathcal {S}$. To ensure that pre-trained models are not biased towards $\mathcal{D}_{S}$, the number of identities of $\mathcal S$ is not larger than the one of $\mathcal {R}$. In specific, great performance is achieved when the number of identities of $\mathcal S$ is set to 100. We simply select the first 100 IDs of $\mathcal{D}_{S}$.

\textbf{Fine-tuning Stage.} To further improve the performance on $\mathcal{D}_{R}$, we fine-tune the pre-trained model without $\mathcal {S}$. Although there exists great domain bias between $\mathcal{D}_{R}$ and $\mathcal{D}_{S}$, low-level features such as color and texture are shared in these two domains. Therefore, first two layers of the pre-trained model are frozen to keep low-level features in the fine-tuning stage. Reducing the learning rate is also necessary.

\subsection{Identity Mining}

The testing set is allowed to be used for unsupervised learning. A widely used method is to use the clustering method to label the data with pseudo labels. Due that the testing set includes 333 identities/categories, we can directly use the k-means clustering to cluster the testing data into 333 categories. However, this way dose not work in Track2 because the poor model can not give accurate pseudo labels. When adding these automatically annotated data to train the model, we observe that the performance becomes worse. We argue that it is unnecessary to add all testing data to train the model, but it is necessary to ensure the correctness of pseudo labels. Therefore, we proposed an Identity Mining (IM) method to solve this problem.

The query set is denoted as $Q = \{q_1,q_2,q_3,...,q_m\}$, while the gallery set is denoted as $G = \{g_1,g_2,g_3,...,g_n\}$. We use the model trained by MDL to extract global features of $Q$ and $G$, which are denoted as $f_Q = \{f_{q_1},f_{q_2},f_{q_3},...,f_{q_m}\}$ and $f_G = \{f_{g_1},f_{g_2},f_{g_3},...,f_{g_n}\}$, respectively. As shown in Figure \ref{fig:im}, the first stage is to find samples of different IDs to form the set $L=\{l_1,l_2,l_3,...,l_t \}$ from $Q$. We randomly sample a probe image $l_1$ to initial set $L$:
\begin{equation}
L=\{l_1\}, l_1 \in Q 
\end{equation}
Then we calculate the distance matrix $Dist(Q,L)$ and define the distance threshold of the negative pair as $d_n$. The goal is to find a sample of new ID to add into set $L$. To achieve such goal, $q_i$ is considered as a candidate when the sub-matrix $\min(Dist(q_i,L))>d_n$. However, there may be multiple candidates satisfying this constraint. We select the most dissimilar candidate with all samples in set $L$ as follow:
\begin{align*}
&l_t =  \mathop{\arg\max}_{q_i}  \sum Dist(q_i,L), \quad q_i \in Q\\
& s.t. \quad \min(Dist(q_i,L)) > d_n\\
\end{align*}
where $l_t$ will be added into the set $L$ with a new ID. We will repeat the process until no $q_i$ satisfies the constraint $\min(Dist(q_i,L))>d_n$. 

After the first step, the set $L$ contains several samples with different IDs. The second step is mining samples belonging to same IDs. Similarly, we define the distance threshold of the positive pair as $d_p$. For an anchor image $l_t$, if a sample $x,x\in Q\cup G$ satisfies $Dist(x,l_t) < d_p$, the sample $x$ will be labeled with the same ID with $l_t$. However, $x$ can be labeled with multiple IDs under this constraint. A simple solution is to label $x$ and the most similar $l_t$ as the same ID. Then, these samples are added into set $L$ with pseudo labels. It is noted that only a part of samples are labeled because we set $d_p < d_n$.  

Compared with the k-means clustering, our IM method, which can automatically generate the clustering centers in the first stage, does not need to know the number of classes. However, the proposed method is a local optimization that is sensitive to the initial sample of $L$. In the future, we will further study it as a global optimization problem. We consider it has the potential to obtain better pseudo labels than other clustering methods.

\subsection{Tracklet-Level Re-Ranking with Weighted Features}

The tracklet IDs are provided for the testing set in Track2. A prior knowledge is that all frames of a tracklet belong to the same ID. In the inference stage, standard ReID task is an image-to-image (I2I) problem. However, with the tracklet information, the task becomes an image-to-track (I2T) problem. For the I2T problem, the feature of tracklet is represented by features of all frames of the tracklet.

He \emph{et al.} \cite{he2019multi} compared average features (AF) and weighted features (WF) of tracklets. Specifically, for a tracklet $T_i = {t_{i,1},t_{i,2},t_{i,3},...,t_{i,j}}$, the average feature is calculated as:
\begin{equation}
f_{T_i} = \frac{1}{j} \sum f_{t_{i,j}}
\end{equation}
However, although some frames are of poor quality due to occlusion, viewpoint or illumination, average features give each frame the same importance. Therefore, average features remains some challenge in I2T ReID. 

He \emph{et al.} \cite{he2019multi} proposed weighted features to address this problem. Specifically, at first, we calculate the sub-matrix $Dist(Q, T_i)$, where $T$ is the set of tracklet from gallery, and $T_i$ is the $i^{th}$ trajectory in $T$. Then we choose the rows of
$Dist(Q, Ti)$ whose min values are lower than 0.2, denoted as $D^{'}$, to get the most similar images as the trajectory. Then We calculate the mean value of each column in $D^{'}$ to get an average distance vector $A_i$ of $T_i$. The weights can
be calculated as follow:
\begin{equation}
W_{i j}=\frac{1}{A_{i j}+0.01}
\end{equation}
Then, the weighted feature of the tracklet $T_i$ can be calculated as follow:
\begin{equation}
f_{T_{i}}=\sum_{j=0}^{s_{i}} F\left(T_{i}\right)_{j} * W_{j}
\end{equation}
, where $f_{T_{i}}$ is the weighted features of $T_i$, and $F(T_i) = \{f_{t_{i,1}},f_{t_{i,2}},f_{t_{i,3}},...,f_{t_{i,j}} \}$ is
the feature set of $T_i$. Then we can get the image-to-track distance matrix $Dist(Q, T)$ to get the ReID result and tile tracks’ images.

Apart from weighted features, $k$-reciprocal re-ranking (RR) method is another post-processing method that can improve ReID performance. However, RR is a frame-level method for image-to-image ReID. To address such problem, we regard each probe image as an independent tracklet which has one frame. Then RR method can be applied into features of tracklet. We name it as track-level re-ranking with weighted features (WF-TRR).
\section{Experimental Results}

\subsection{Datasets}
Different from the standard vehicle ReID task defined by the academic researches, the Track2 can train models on both real-world and synthetic data, as shown in Figure \ref{fig:demo}. In addition, unsupervised learning on the testing set is also permitted.

\textbf{Real-world data. }The real-world data, which is called as CityFlow dataset\cite{Naphade19AIC19,Tang19CityFlow} in this paper, is captured by 40 cameras in real-world traffic surveillance environment. It totally includes 56277 images of 666 vehicles. 36935 images of 333 vehicles are used for training. The remaining 18290 images of 333 vehicles are for testing. In the testing set, there are 1052 query images and 17238 gallery images, respectively. On average, each vehicle has 84.50 image signatures from 4.55 camera views. The single-camera tracklets are provided on both training and testing sets. The performance on the testing set determines the final ranking on the leaderboard.

\textbf{Synthetic data. }The synthetic data, which is called as VehicleX dataset in this paper, is generated by a publicly available 3D engine VehicleX \cite{Yao19VehicleX}. The dataset only provides training set, which contains 192150 images of 1362 vehicles in total. In addition, the attribute labels, such as car colors and car types, are also annotated. In Track2, the synthetic data can be used for the model training or transfer learning. However, there exists great domain bias between the real-world and synthetic data.

\textbf{Validation data. }Since each team has only 20 submissions, it is necessary to use the validation set to evaluate methods offline. We split the training set of CityFlow into the training set and the validation set. For convenience, we named them as Split-train and Split-test, respectively. Split-train and Split-test include 26272 images of 233 vehicles and 10663 images of 100 vehicles, respectively. In Split-test, 3 images are sampled as probe for each vehicle, and the remaining images are regarded as gallery.

\subsection{Implement Details}
All the images are resized to 320 $\times$ 320.  As shown in the Figure \ref{fig:baseline}, We adopt the ResNet101$\_$IBN$\_$a \cite{pan2018IBN-Net} as the backbone network. As for data augmentation, we use random flipping, random padding and random erasing. In the training stage, we use soft margin triplet loss with the mini-batch of 8 identities and 12 images of each identity which leads to better convergence. SGD is applied as the optimizer and the initial learning rate is set to $1e^{-2}$. Besides, we adopt the Warmup learning strategy and spend 10 epochs linearly increasing the learning rate from $1e^{-3}$ to $1e^{-2}$. The learning rate is decayed to $1e^{-3}$ and $1e^{-4}$ at 40th and 70th epoch, respectively. We totally train the model with 100 epoches.

For the pre-trained stage of MDL, 14536 images of first 100 identities in VehicleX are added to pre-train the model. These sampled images are fixed at each epoch. For the fune-tuning stage of MDL, the model is fine-tuned with 40 epoches, where the initial learning rate is set to $1e^{-3}$ for the layer in the backbone network and $2e^{-3}$ for the fully-connect layer. For the IM method, distance threshold $d_n$ is set to 0.49 and $d_p$ is set to 0.23. The features used to calculate the distance are normalized by L2 normalization. In total, 7084 images of 130 identities were selected as set $L$ from 333 categories. 

\subsection{Comparison of Different Re-Ranking Strategies}

\renewcommand{\multirowsetup}{\centering}
\begin{table}[htb]
  \begin{center}
  \begin{tabular}{ c|cc|cc}
\hline
    		& \multicolumn{2}{c|}{Split-test} & \multicolumn{2}{c}{CityFlow}	 \\
    Model	 & mAP  & r = 1 	& mAP	&r = 1 	 	 \\
 	\hline
	\hline
    Baseline(BoT-BS)   	&76.7 &89.7	&/	&/ \\
    BoT-BS + RR  \cite{zhong2017re} &80.4	&89.3 &54.5	&67.4		\\
    BoT-BS + AF-TRR     &82.0  & 90.1  &55.5  &67.3  \\
    BoT-BS + WF-TRR \cite{he2019multi} 	&\textbf{90.6} &\textbf{93.0}		&\textbf{59.7} &\textbf{69.3}	\\
\hline
  \end{tabular}
  \end{center}
  \caption{\label{tab:trr}Comparison of different re-ranking strategies. RR and TRR mean k-reciprocal Re-Ranking and Track-Level Re-Ranking. AF and WF mean that average feature and weighted feature of each tracklet. BoT-BS is not evaluated on CityFlow.}
\end{table}

We evaluate different re-ranking strategies on the validation set Split-test in Table \ref{tab:trr}. The baseline BoT-BS achieves 76.7\% mAP on Split-test. The image-to-image k-reciprocal re-ranking (RR) \cite{zhong2017re} improves 3.7\% mAP on Split-test. Since tracklet IDs are provided, we evalute two tracklet-level re-ranking strategies (TRR), i.e., AF-TRR and WF-TRR. Directly averaging all features of the tracklet to represent its features, AF-TRR beats RR by 1.6\% mAP on Split-test. It demonstrates that the tracklet information is useful for the ReID performance. Finally, WF-TRR obtains significant performance reaching 90.6\% mAP on Split-test, surpassing RR and AF-TRR by large margins.

On CityFlow benchmark, RR, AF-TRR and WF-TRR achieve 54.5\%, 55.5\%, 59.7\% mAP, respectively. The weighted features weaken the contribution of some low-quality images. BoT-BS + WF-TRR is the baseline for the following sections.

\subsection{Analysis of Multi-Domain Learning}

\renewcommand{\multirowsetup}{\centering}
\begin{table}[htb]
  \begin{center}
  \begin{tabular}{ c|cc}
\hline
    	 & \multicolumn{2}{c}{Split-test}	 \\
    Model	   & mAP 	& r = 1	 	 \\
 	\hline
	\hline
    Baseline(BoT-BS)    &76.7 &89.7     \\
    Solution-1   &72.9	& 85.7	\\
    Solution-2   &71.5	& 83.0	\\
    BoT-BS + MDL(50 IDs)   &77.1 &88.7     \\
    BoT-BS + MDL(100 IDs)   &79.4 &90.7     \\
    BoT-BS + MDL(200 IDs)   &76.6 &88.3     \\
    BoT-BS + MDL(300 IDs)   &76.8 &87.0     \\
\hline
  \end{tabular}
  \end{center}
  \caption{\label{tab:mdl-1}The results of MDL on Split-test.}
\end{table}

\renewcommand{\multirowsetup}{\centering}
\begin{table}[htb]
  \begin{center}
  \begin{tabular}{ c|cc}
\hline
    	 & \multicolumn{2}{c}{CityFlow}	 \\
    Model	   & mAP 	& r = 1		 \\
 	\hline
	\hline
    Baseline(BoT-BS+WF-TRR)   & 59.7	& 69.3     \\
    Baseline + MDL &65.3	& 72.0			\\
\hline
  \end{tabular}
  \end{center}
  \caption{\label{tab:mdl-2}The results of MDL on CityFlow.}
\end{table}

We evaluate multi-domain learning (MDL) method on Split-test and CityFlow. In Table \ref{tab:mdl-1}, both Solution-1 and Solution-2 cannot surpass the performance of BoT-BS. Because there exists great bias between the real-world data and the synthetic data, joint training or pre-training are invalid for Track2. In addition, we evaluate our MDL method with adding different number of IDs. The results in Table \ref{tab:mdl-1} show that MDL (100 IDs) achieves the best performance. If the first two layers are not frozen during the fine-tuning stage, the performance will be reduced by 1.4\% mAP. Less identities provide less knowledge while more identities may cause the model to be more biased towards the synthetic data. However, MDL achieves better performance than Baseline, Solution-1 and Solution-2, which shows its effectiveness.

On CityFlow benchmark, MDL also improves the mAP accuracy from 59.3\% to 65.3\% in Table \ref{tab:mdl-2}.

\begin{figure*}[htb]
\centering
\includegraphics[width=1.\linewidth]{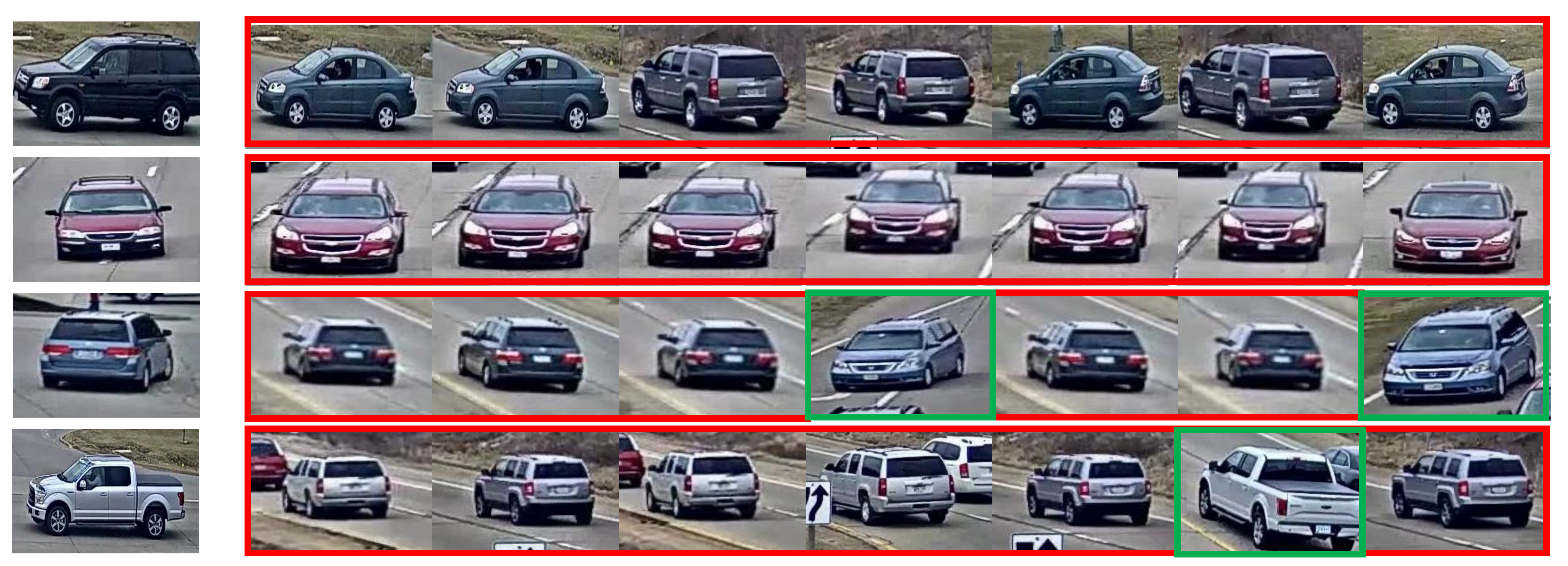}
\caption{Visualization results of the baseline model (BoT-BS). The first column shows several different query images, each row demonstrates the 10-nearest gallery images ranked by the model. Green and red box correspond to the positives and negatives, respectively.}
\label{fig:result1}
\end{figure*}

\begin{figure*}[htb]
\centering
\includegraphics[width=1.\linewidth]{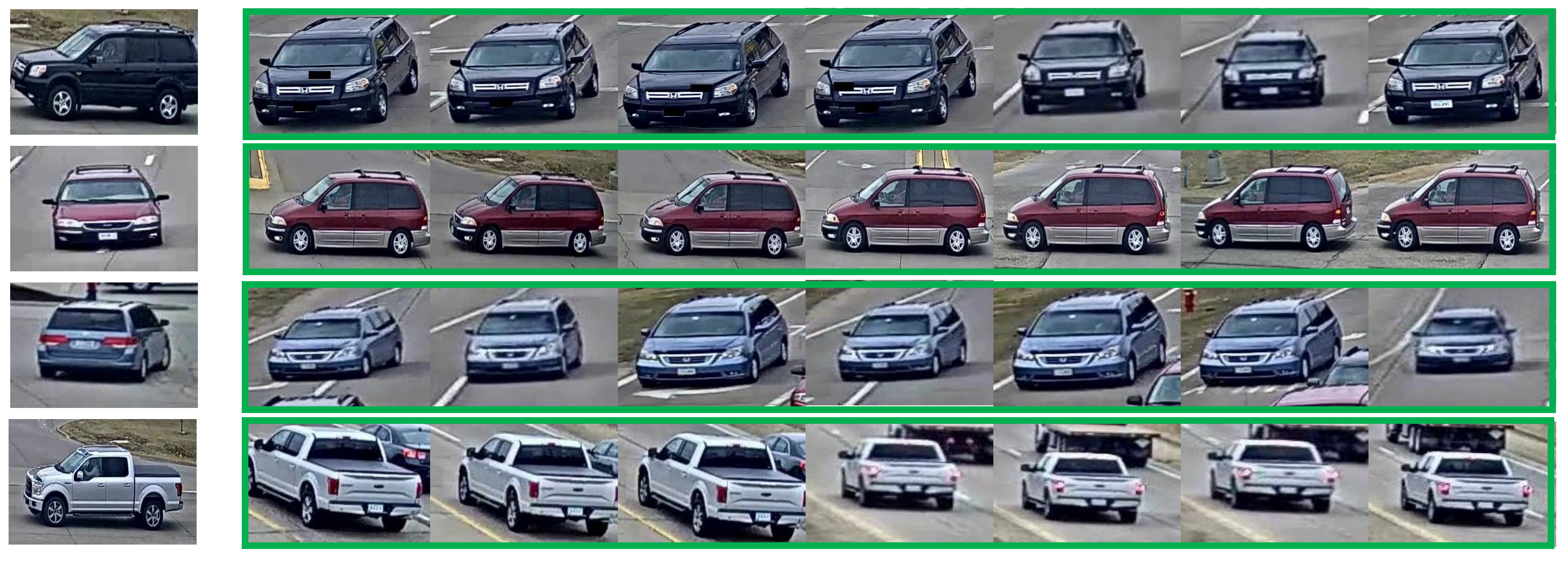}
\caption{Visualization results of the proposed method. The query images are the same with Figure \ref{fig:result1}.}
\label{fig:result2}
\end{figure*}

\subsection{Analysis of Identity Mining}

We compare results of our IM method and the k-means clustering in Table \ref{tab:im}. For the k-means clustering, the number of categories are set to 100 and 333 for Split-test and CityFlow, respectively. Because of the poor performance of the trained model, k-means clustering cannot generate pseudo labels that are accurate enough. On Split-test, we observe that the accuracy of pseudo labels generated by the k-means clustering is 84.0\%, while our IM method can generate pseudo labels with the accuracy of 98.7\%. The mAP scores on CityFlow further show the effectiveness of our IM method. The k-means clustering cannot improve the ReID performance of Baseline, while Baseline+IM achieves 68.5\% mAP that is better than Baseline by 3.2\% mAP.

\renewcommand{\multirowsetup}{\centering}
\begin{table}[htb]
  \begin{center}
  \begin{tabular}{ c|cccc}
\hline
    	 & \multicolumn{4}{c}{CityFlow}	 \\
    Model	   & mAP 	& r = 1	&r = 5 	& r = 10 	 \\
 	\hline
	\hline
    Baseline(BoT-BS+MDL)   &65.3	& 72.0	& 72.4	& 73.3 \\
    Baseline +k-means &63.9	& 72.2	& 72.2	& 72.6		\\
    Baseline +IM &68.5	& 74.7	& 74.9	& 75.3		\\
\hline
  \end{tabular}
  \end{center}
  \caption{\label{tab:im}The results of IM on CityFlow.}
\end{table}

\subsection{Ablation study on CityFlow}

As shown in Table \ref{tab:city}, BoT-BS with WF-TRR achieves 59.7 mAP on CityFlow, which is improved to 65.3\% by MDL. When jointing the a part of testing data labeled by IM to train the model, the single model with ResNet101-IBN-a backbone reaches 68.5\% on CityFlow. To achieves better performance, we fuse the results of five models reaching 73.2\% mAP. The increase in mAP score shows the effectiveness of each module.

\renewcommand{\multirowsetup}{\centering}
\begin{table}[htb]
  \begin{center}
  \begin{tabular}{ cccc|c}
\hline
    WF-TRR	   & MDL 	& IM	&ENS & mAP	 \\
 	\hline
	\hline
    \checkmark   & 	&     &	& 59.7 \\
    \checkmark   & \checkmark	&     & &65.3		\\
    \checkmark   & \checkmark	&  \checkmark   & &68.5		\\
    \checkmark   & \checkmark	&   \checkmark  & \checkmark&73.2		\\
\hline
  \end{tabular}
  \end{center}
  \caption{\label{tab:city}Ablation study on CityFlow. WF-TRR means tracklet-level re-ranking strategy with weighted features. MDL and IM mean the proposed multi-domain learning method and Identity Mining method, respectively. Ens means the ensemble of five different model.}
\end{table}
\subsection{Competition Results}

Our team (Team ID 39) achieves 0.7322 in the mAP score which yields third place among the total 41 submissions in the 2020 NVIDIA AI City Challenge Track 2. As shown in the Table 2, it is the performance of top-10 algorithms and detailed statistics is shown in the Table \ref{tab:leaderboard}. Some other scores such as rank-1, rank-5 and rank-10 are shown in Table \ref{tab:our}. The scores has not been checked by the organizing committee, so this may not be the final ranking.

\begin{table}[htb]
  \begin{center}
  \begin{tabular}{c|c|c|c}
\hline
    Rank &Team ID	& Team Name & mAP Scores 	 \\
 	\hline
	\hline
    1	&73	&Baidu-UTS	&0.8413 \\
    2	&42	&RuiYanAI	&0.7810 \\
   \textbf{3} &\textbf{39}	&\textbf{DMT(Ours)}	&\textbf{0.7322} \\
    4	&36	&IOSB-VeRi	&0.6899 \\
    5	&30	&BestImage	&0.6684 \\
    6	&44	&BeBetter	&0.6683\\
    7	&72	&UMD\_RC	&0.6668\\
    8	&7	&Ainnovation	&0.6561\\
    9	&46	&NMB	&0.6206\\
    10	&81	&Shahe	&0.6191\\
    \hline
  \end{tabular}
  \end{center}
  \caption{\label{tab:leaderboard}Competition results of AICITY20 Track2.}
\end{table}

\begin{table}[htb]
  \begin{center}
  \begin{tabular}{c|c|c|c|c|c|c|c}
\hline
    mAP & r=1 & r=5 &r=10 &r=15 &r=20 	 \\
 	\hline
	\hline
    0.7322&0.8042	&0.8051	&0.8108	&0.8203	&0.8337 \\
    \hline
  \end{tabular}
  \end{center}
  \caption{\label{tab:our}mAP and CMC scores of our method.}
\end{table}

\section{Conclusion}

The paper introduces the solution of our team (Team ID 39) in the 2020 NVIDIA AI City Challenge Track 2, CVPR2020 Conference. Our solution is based on a strong baseline in person ReID. In addition, we observe that tracklet-level re-ranking strategy improves the ReID performance by post-processing the results. We propose multi-domain Learning method to exploit the synthetic data and Identity Mining method to automatically labeled a part of the testing data. The model can achieve better performance with these additional data. Finally, our team achieves 0.7322 in the mAP score which yields third place among the total 41 submissions.

{\small
\bibliographystyle{ieee_fullname}
\bibliography{egbib}
}

\end{document}